\documentclass[journal]{IEEEtran}

\usepackage{cite}

\ifCLASSINFOpdf

\else

\fi

\usepackage{times}  % DO NOT CHANGE THIS
\usepackage{helvet} % DO NOT CHANGE THIS
\usepackage{courier}  % DO NOT CHANGE THIS
\usepackage[hyphens]{url}  % DO NOT CHANGE THIS
\usepackage{graphicx} % DO NOT CHANGE THIS
\usepackage{graphicx}  % DO NOT CHANGE THIS
\usepackage{amsmath,graphicx}

\usepackage{amsmath}
\usepackage{xcolor,soul,framed} %,caption
\usepackage{xspace}
\usepackage{array}
\usepackage{eqparbox}
\usepackage{url}
\usepackage{longtable}
\usepackage{lipsum}
\usepackage{blindtext}
\usepackage{makecell}
\usepackage{mathtools}
\usepackage{commath}
\usepackage{multirow}
\usepackage{tabularx}
\usepackage[normalem]{ulem}
\usepackage{xspace}
\usepackage{algorithm}
\usepackage{algpseudocode}
\usepackage{amsfonts}
\usepackage{placeins}

\usepackage[normalem]{ulem}

\newcolumntype{C}{>{\hsize=\dimexpr0.5\hsize+8\tabcolsep+\arrayrulewidth\centering\relax}X}
\DeclareMathOperator{\cref}{ref}

\begin{document}

\title{Capsule Attention for Multimodal EEG-EOG Representation Learning with Application to Driver Vigilance Estimation}% in Smart Vehicles}

\author{Guangyi~Zhang 
        and~Ali~Etemad,~\IEEEmembership{Senior Member,~IEEE}% <-this % stops a space
\thanks{G. Zhang, and A. Etemad are with the Department of Electrical and Computer Engineering, Queen's University, Kingston, ON, Canada (e-mail: guangyi.zhang@queensu.ca,ali.etemad@queensu.ca).}

}

\maketitle

\begin{abstract}
Driver vigilance estimation is an important task for transportation safety. Wearable and portable brain-computer interface devices provide a powerful means for real-time monitoring of the vigilance level of drivers to help with avoiding distracted or impaired driving. In this paper, we propose a novel multimodal architecture for in-vehicle vigilance estimation from Electroencephalogram and Electrooculogram. To enable the system to focus on the most salient parts of the learned multimodal representations, we propose an architecture composed of a capsule attention mechanism following a deep Long Short-Term Memory (LSTM) network. Our model learns hierarchical dependencies in the data through the LSTM and capsule feature representation layers. To better explore the discriminative ability of the learned representations, we study the effect of the proposed capsule attention mechanism including the number of dynamic routing iterations as well as other parameters. Experiments show the robustness of our method by outperforming other solutions and baseline techniques, setting a new state-of-the-art. We then provide an analysis on different frequency bands and brain regions to evaluate their suitability for driver vigilance estimation. Lastly, an analysis on the role of capsule attention, multimodality, and robustness to noise is performed, highlighting the advantages of our approach.
\end{abstract}

\section{Introduction} \label{sec:intro}
Recent advances in driver monitoring using modern sensing technologies have the potential to reduce the number of driving accidents, especially those occurring due to driver fatigue, distraction, and the influence of illegal substances. Accordingly, recent studies have tackled the notion of measuring and monitoring driver awareness, also referred to as \textit{vigilance} \cite{zheng2017multimodal}. For example, in recent years, wireless and wearable devices have been used to collect signals such as Electroencephalogram (EEG) and Electrooculogram (EOG) for estimation of driver alertness \cite{lin2014wireless,larue2011driving,ma2014eog}. Recently, deep learning methods are being increasingly used for EEG analysis \cite{zhang2019classification,zhang2020rfnet}. While monitoring systems built by means of advanced deep learning techniques generally require considerable resources during the training phase, deployment is often not limited by computational power as inference is achievable in real-time \cite{shawki2020deep}.

In general, EEG, which captures brain activity recorded from the scalp, is influenced by factors such as fatigue and alertness during different activities such as driving \cite{wang2015real}. Similarly, EOG which collects the potentials between the front and back of human eyes, notably cornea and retina, and is recorded from the forehead \cite{ma2014eog}, contains information regarding vigilance and eye movements (e.g., blinking and saccade) \cite{galley1993evaluation}. The fusion of EEG and EOG (multimodal) has subsequently been utilized for analysis of vigilance, showing clear advantages over EEG and EOG alone \cite{zheng2017multimodal,du2017detecting,huo2016driving,wu2018regression,zhang2016continuous}. 
However, multimodal learning of EEG and EOG for in-vehicle vigilance estimation remains a challenging task due to a number of open problems. First, much like other biological signals, EEG and EOG are often contaminated by environmental artifacts and noise. Moreover, EEG and EOG are susceptible to artifacts caused by motion and muscle activity such as jaw motion, frowning, and others, making their interpretation particularly challenging \cite{chaumon2015practical,tang2020unified}. Lastly, multimodal analysis of biological signals is generally difficult since identifying the complementary and contradicting information in the available signals is a challenging \cite{wu2018regression}.

\begin{figure*}[!t]
    \begin{center}
    \includegraphics[width=0.75\linewidth]{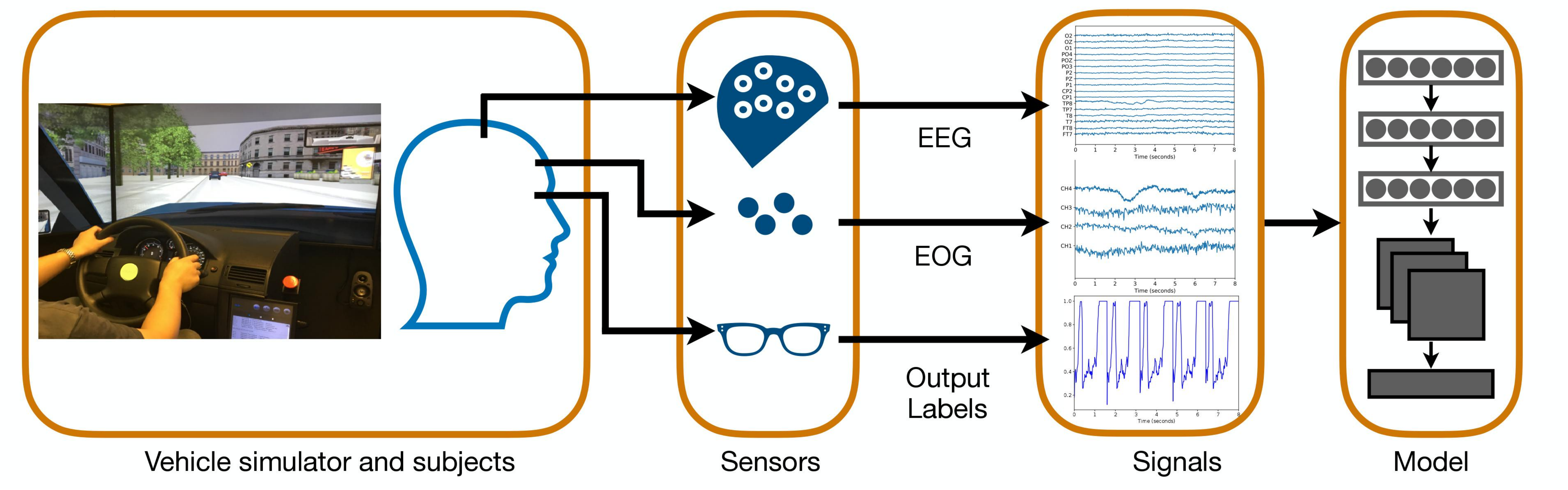} 
    \end{center}
\caption{The overview of the experiment work-flow is presented.}
\label{fig-overview}
\end{figure*}

We believe the solution to the problems mentioned above lies in an architecture capable of learning the temporal relationships followed by the ability to focus on certain sections within the learned representations in order to selectively attend to different parts of the data given the redundant, complementary, uncertain, or noisy information. As a result, in this paper, in order to perform driver monitoring through vigilance estimation, we propose a novel solution that first encodes multimodal EEG-EOG information through a deep LSTM network, and then learns the hierarchical dependencies and part-whole relationships in the learned representations through a \textit{capsule attention} mechanism. Specifically, the capsule attention mechanism establishes a hierarchical relationship between lower level capsules (containing part information) and higher level capsules (containing whole information). As a result, this capsule-based mechanism guides higher level capsules to capture features that are more robust to input noise and artifacts, thus providing more accurate predictions of vigilance. We compare our proposed model to a number of other works including past published methods and our own baselines. We illustrate that our model significantly outperforms the state-of-the-art solutions in both intra-participant and cross-participant validation schemes, with lower Root Mean Square Error (RMSE) and higher Pearson Correlation Coefficient (PCC). An overview of the system is illustrated in Figure \ref{fig-overview}. 

Our contributions in this paper can be summarized as follows. (\textbf{1}) For the first time, we propose the use of a Capsule-based Attention mechanism for multimodal EEG and EOG representation learning. Our model uses a deep LSTM network followed by lower and higher -level capsules to adaptively focus on the salient time-steps and modalities in order to learn effective representations prior to classification. 
(\textbf{2}) Our analysis shows that our model is capable of learning representations that are much more separable compared to the original feature space for unimodal EEG, unimodal EOG, and multimodal EEG-EOG. (\textbf{3}) We perform detailed experiments on the robustness of our method with respect to added noise with different intensity levels. The analysis demonstrates that our proposed method is more robust to noise compared to other baselines.

The rest of this paper is organized as follows. Section II provides an overview of related work on EEG-based, EOG-based and multimodal EEG-EOG vigilance estimation as well as soft attention and capsule attention. In Section III, we give a systematic description of the proposed architecture including problem setup, solution overview, and architecture details. Section IV gives a description of the dataset we used, implementation details, evaluation methods, and comparisons between our model and other existing pipelines. In section V, we present the results, perform ablation experiments, and provide further analysis. In Section VI, we present the summary as well as conclusions of our work.

\section{Related Work}
In this section, we review the related work on vigilance estimation. First, we discuss prior works that have only used EEG for this purpose, followed by approaches that have only used EOG for the same task. This is followed by a review of multi-modal EEG-EOG solutions. In the end, we review an important concept that is highly relevant to our proposed solution, namely capsule attention.

\subsubsection{Unimodal EEG Vigilance Estimation}
EEG has been commonly used as an effective measurement for estimating vigilance \cite{shi2013eeg,ko2017sustained}. Most of the existing pipelines consist of feature extraction followed by a supervised classifier for learning the relationship between the extracted features and output vigilance scores \cite{shi2013eeg,zheng2017multimodal}. For example, in \cite{shi2013eeg}, the logarithm of the power spectrum representation of EEG was adopted as a powerful feature, while Support Vector Machine (SVM) and Extreme Learning Machine (ELM) were used as classifiers, achieving considerable accuracy in continuous vigilance level estimation. Artificial Neural networks (ANN) have also been used to recognize different vigilance states such as awake, drowsy, and sleep using EEG and Electromyogram (EMG) as input \cite{akin2008estimating}. To reduce the high dimensionality of EEG features, Principle Component Analysis (PCA) has been used directly on the frequency domain representations of EEG in order to increase the performance of vigilance estimation \cite{shi2013robust,cao2010eeg}. To explore sources from mixed multi-channel EEG, Independent Component Analysis has been applied on EEG data. Several classifiers such as SVM, Gaussian Classifier (GC), and Radial Basis Function Neural Network (RBFNN) have been employed to achieve promising results in recognizing cognitive states \cite{chuang2013independent,chuang2015eeg}. Moreover, Common Spatial Patterns (CSP) followed by PCA have been used as a feature extraction approach instead of directly using features from the EEG spectrum \cite{shi2008dynamic}. In particular, CSP has been used as a spatial filter while PCA helps to select the most relevant spectral features based on the highest eigenvalues. Unsupervised algorithms have also been used based on the distribution of extracted EEG features. For example, a dynamic clustering algorithm has been utilized to build relationships between EEG features and several vigilance categories \cite{shi2008dynamic}. Furthermore, Convolutional Neural Network (CNN) has been employed to extract features from raw EEG signals for estimating driver fatigue \cite{gao2019eeg,gao2020complex}.

\subsubsection{Unimodal EOG Vigilance Estimation}
EOG signals record information regarding eye movements. Generally, EOG signals contain higher signal-to-noise ratios compared to EEG \cite{zheng2019vigilance}. In order to investigate the correlation between EOG and vigilance, several algorithms have been developed to extract features of eye movements such as slow eye movements (SEM), saccade, and blinks from EOG signals \cite{ma2014eog}. In \cite{ma2010vigilance}, SEM outperformed rapid eye movement and blink in vigilance estimation. Combination of these eye movement features after Linear Dynamic System (LDS) processing reached a high correlation with vigilance in \cite{ma2014eog}. Other than manual feature extraction, a CNN was applied in \cite{zhu2014eog} on raw EOG signals to automatically extract representations that result in a good performance when detecting drowsiness.

\newcolumntype{T}{>{\centering\arraybackslash}X}
\begin{table*}[!ht]
\centering

\caption{A summary of the related work and their main characteristics.}\label{tab:related work}

\scriptsize
\setlength\tabcolsep{5.0pt}
\begin{tabularx}{\textwidth}{cccTcTcTT}
	\hline
	\multirow{2}{*}{Ref.}  & \multirow{2}{*}{Year} & Method  & \multirow{2}{*}{Method}    & \multirow{2}{*}{Modalities}        & \multirow{2}{*}{Task}       & \multirow{2}{*}{Dataset} & Validation Protocol & Validation             \\
	&    & {Type}  & &&  &      & (No. of Participants) & Scheme                \\
	\hline \hline
	
	\cite{akin2008estimating}           & 2008     & Deep             & ANN                        & EEG, EMG               & Classification             & N/A               & Intra-Participant ($30$)       & $50:50$ Split                   \\
	\cite{cao2010eeg}                   & 2010     & Classical        & SVM                        & EEG                    & Classification             & N/A               & Intra-Participant ($9$)        & $50:50$ Split                   \\
	\cite{ma2010vigilance}              & 2010     & Classical        & LDS                        & EOG                    & Regression                 & N/A               & Intra-Participant ($5$)        & Pre-defined                     \\        
	\cite{shi2013robust}                & 2013     & Classical        & SVR                        & EEG                    & Regression                 & N/A               & Intra-Participant ($23$)       & Pre-defined                     \\
    \cite{shi2013eeg}                   & 2013     & Classical        & ELM                        & EEG                    & Regression                 & BCMI              & Intra-Participant ($9$)        & Pre-defined                     \\
	\cite{chuang2013independent}        & 2013     & Classical        & SVM, GC, RBFNN             & EEG                    & Classification             & N/A               & Cross-Participant ($10$)       & LOSO                            \\
	\cite{ma2014eog}                    & 2014     & Classical        & LDS                        & EOG                    & Regression                 & N/A               & Intra-Participant ($22$)       & Pre-defined                     \\
	\cite{zhu2014eog}                   & 2014     & Deep             & CNN                        & EOG                    & Regression                 & N/A               & Cross-Participant ($22$)       & LOSO                            \\
	\cite{huo2016driving}               & 2016     & Classical        & GELM                       & EEG, EOG               & Regression                 & SEED-VIG          & Intra-Participant ($21$)       & $5-$Fold                        \\   
	\cite{zhang2016continuous}          & 2016     & Deep             & LSTM                       & EEG, EOG               & Regression                 & SEED-VIG          & Intra-Participant ($21$)       & $5-$Fold                        \\  
	\cite{zheng2017multimodal}          & 2017     & Classical        & SVR, CCRF, CCNF            & EEG, EOG               & Regression                 & SEED-VIG          & Intra-Participant ($23$)       & $5-$Fold                        \\
    \cite{du2017detecting}              & 2017     & Deep             & DAE                        & EEG, EOG               & Regression                 & SEED-VIG          & Intra-Participant ($21$)       & $5-$Fold                        \\	
	\cite{li2018multimodal}             & 2018     & Deep             & DANN, ADDA                 & EEG, EOG               & Regression                 & SEED-VIG          & Cross-Participant ($23$)       & LOSO                            \\
	\cite{wu2018regression}             & 2018     & Deep             & DNNSN                      & EEG, EOG               & Regression                 & SEED-VIG          & Intra-Participant ($23$)       & $5-$Fold                        \\   
	\cite{zheng2019vigilance}           & 2019     & Classical        & SVR, CCRF, CCNF            & EOG                    & Regression                 & SEED-EOG          & Intra-Participant ($20$)       & $5-$Fold                        \\   	
	\cite{gao2019eeg}                   & 2019     & Deep             & CNN                        & EEG                    & Classification             & N/A               & Intra-Participant ($8$)        & $10-$Fold                       \\
	\cite{ko2020eyeblink}               & 2020     & Classical        & MLR                        & EEG, EOG               & Regression                 & N/A               & Cross-Participant ($15$)       & LOSO                            \\   
    \hline% \hline
\end{tabularx}
\end{table*}

\subsubsection{Multimodal EEG-EOG Vigilance Estimation}
A number of prior works have focused on analyzing the added value of using both EEG and EOG (multimodal) as opposed to corresponding unimodal methods. For example, Support Vector Regression (SVR) was employed for unimodal EEG, unimodal EOG, and multimodal EEG-EOG respectively, demonstrating that EEG and EOG have \textit{complementary information} for vigilance estimation \cite{zheng2017multimodal}. The superiority of multimodal vigilance estimation was also confirmed in \cite{huo2016driving} using Graph-regularized Extreme Learning Machine (GELM), achieving better performance with multimodal EEG and EOG compared to individual EEG and EOG.

Several conventional machine learning solutions have been proposed for driving vigilance evaluation with multimodal EEG and EOG data. For instance, in \cite{zheng2017multimodal}, two probabilistic models notably Continuous Conditional Random Field (CCRF) and Continuous Conditional Neural Field (CCNF) were employed for multimodal vigilance estimation. Moreover, a Multiple Linear Regression (MLR) model was used to predict fatigue based on the combination of EEG features and eye blink features \cite{ko2020eyeblink}.

Several deep learning methods have also been used in vigilance estimation. In \cite{du2017detecting}, a multimodal Deep Auto Encoder (DAE) was employed. In \cite{zhang2016continuous}, the use of an LSTM network resulted in a considerable improvement using feature fusion over single-mode EEG and EOG. In \cite{wu2018regression}, a Double-layered Neural Network with Subnetwork Nodes (DNNSN) was utilized along with multimodal feature selection using an autoencoder, obtaining impressive results. In \cite{li2018multimodal}, two domain adaption networks, notably Domain-Adversarial Neural Network (DANN) and Adversarial Discriminative Domain Adaptation (ADDA) were employed with feature fusion.      

We summarize the main works on vigilance estimation and characteristics of the existing pipelines including input modalities and methods in Table \ref{tab:related work}. We also cover information about the used datasets and validation protocols. We sort the related works in the table by the date of publication. According to the table, a few interesting trends can be observed. First, recent works have mostly employed deep learning methods. Moreover, most recent works tend to exploit multimodal EEG-EOG for vigilance estimation. Lastly, we observe that most works in this area evaluate the proposed solutions based on the less challenging `intra-participant' scheme. Similar to most of the recent works in this area, in this paper, we use multimodal EEG-EOG for driver vigilance estimation with deep learning, while utilizing \textit{both} intra-participant and cross-participant protocols for evaluation.

\subsubsection{A Background on Capsule Attention}
% \subsubsection{Soft Attention}
Attention mechanisms have been proposed to enable models such as RNNs and CNNs to focus on salient components within the representations obtained by the previous layers of the pipeline. For example, Soft Attention (SoftAtt) mechanisms were recently proposed for Natural Language Processing (NLP) \cite{wang2016attention}, to be used with LSTM networks, and have since been used for other applications, including EEG analysis \cite{zhang2019classification}. This mechanism results in better feature representation learning by assigning learned weights to LSTM cell outputs.

Capsule networks were proposed in \cite{sabour2017dynamic} in $2017$ and have shown strong characteristics in learning hierarchical relationships in the input data, outperforming other deep learning architectures in a number of applications such as facial expression recognition \cite{hosseini2019gf} and infrared facial image recognition \cite{vinay2018optimal}. These networks were proposed to capture important high-level information by learning part-whole relationships using capsules (group of neurons) with dynamic routing to overcome a number of limitations in CNNs and RNNs \cite{sabour2017dynamic}. Specifically, the lower level capsule contains local features (part information) with fewer degrees of freedom. While the higher level capsule contains global features (whole information) with more degrees of freedom. Such hierarchical relationship represents how much information from each lower level capsule is contributed to a higher level capsule. The higher level capsules are able to learn the variations of perturbation of features (e.g., skew, style, scale, and thickness in the image domain). Moreover, experiments have shown that the capsule network is more robust to small affine transformations of the features compared to traditional CNNs \cite{sabour2017dynamic}. In the EEG domain, the recordings often contain noise and artifacts, for example, caused by electrode placement, hair, etc. We hypothesize that an architecture based on capsule networks can provide a more robust prediction on vigilance using EEG data.
While capsule networks can be used on their own for learning, in this paper, we use it as a form of attention mechanism successive to a deep LSTM network. \textit{Capsule attention} employs routing by agreement to enable the lower level capsules to learn what needs to be paid attention to given the feedback from higher level capsules. Lower level capsules will then route to the higher level capsules by similarity agreement. This concept has been very recently used in state-of-the-art NLP relation extraction \cite{zhang2018attention} and visual question answering \cite{zhou2019dynamic}, and to the best of our knowledge, has not been utilized on biological and wearable representation learning despite its promising initial results in other fields.

\begin{figure*}[!t]
    \begin{center}
    \includegraphics[width=0.8\linewidth]{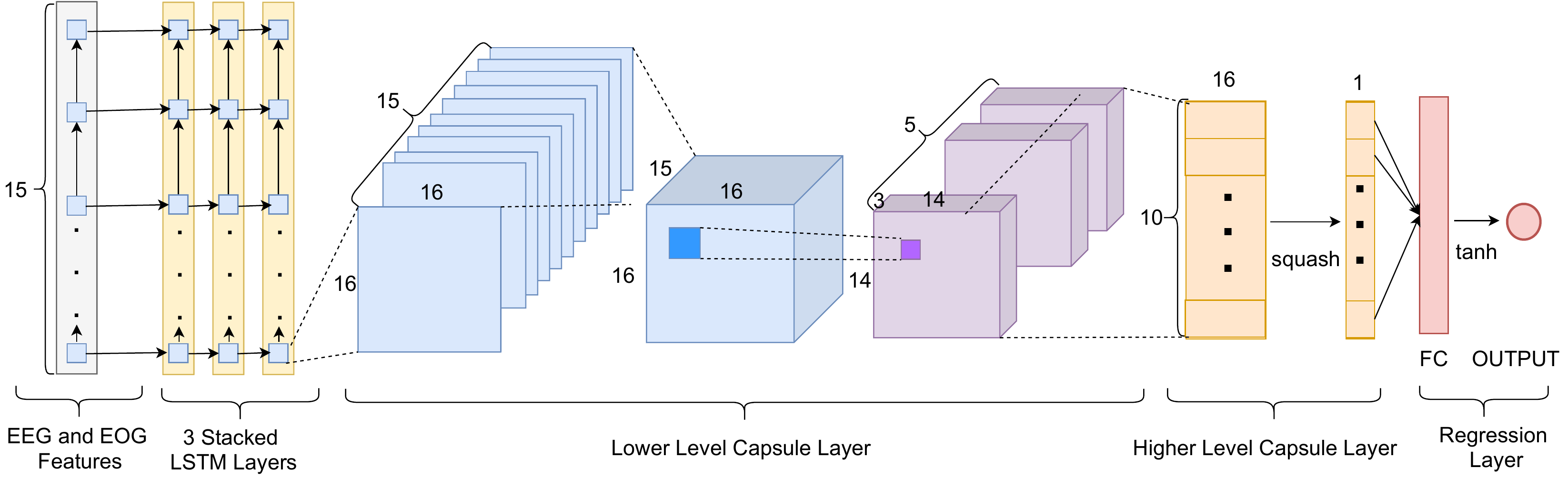} 
    \end{center}
\caption{The architecture of our proposed method is presented.}
\label{fig-arch}
\end{figure*}

\section{Proposed Architecture}
\subsection{Problem Setup}
Suppose $\big\{ (x_i^j, y_i^j) \in  X \times Y : \forall {i}\in [1,r], \forall{j}\in [1,s] \big\}$ denote the set of input data and labels, where $i$ and $j$ denote the sample and participant indices respectively. $r$ is the number of samples belonging to each participant and $s$ is the total number of participants. Due to the biological differences among participants and even the same participant at different times, biological signals especially EEG and EOG, are very participant- and session-dependant \cite{zhang2018cascade}. This phenomenon has resulted in the adoption of distinct intra- and cross-participant validation schemes:

\noindent \textbf{\textit{i}) Intra-participant scheme:} In this validation scheme, we equally split $X \times Y$ into $k$ number of folds. For the $m$th iteration for the $n$th participant ($m \in [1, k], n \in [1,s]$) we set $(X_{train}, Y_{train})= \big\{ (x_{i,l}^j, y_{i,l}^j) \in X \times Y: \forall i \in [1, r], l \neq m, j = n, \forall l \in [1,k] \big\}$, where $l$ is the iteration index, and $(X_{test}, Y_{test})= \big\{ (x_{i,l}^j, y_{i,l}^j) \in X \times Y: \forall{i}\in [1, r], l = m, j = n  \big\}$.

\noindent \textbf{\textit{ii}) Cross-participant scheme:} In this validation scheme, for the $h$th experiment ($h\in [1,s]$), we have $(X_{train}, Y_{train}) = \big\{ (x_i^j, y_i^j) \in X \times Y:  \forall i \in [1, r], j \neq h, \forall j \in [1, s] \big\}$ and $(X_{test}, Y_{test}) = \big\{ (x_i^j, y_i^j) \in X \times Y$: $\forall i \in [1, r], j = h \big\}$.

\subsection{Solution Overview}
We design our model with the aim of learning dependencies and discriminative information from the multimodal data. To achieve this, an LSTM is first used to learn the temporal dependencies in the data. Next, to deal with the inherent challenges in multimodal biological data as described earlier in the Introduction (e.g. complementary or contradictory information, lack of control on participant mental activity, and others), we propose the use of capsule attention to learn the part-whole hierarchical relationships in the representations received from the LSTM outputs. This section describes our model, which consists of five layers, namely, input representation layer, LSTM layer, lower level capsule layer, higher level capsule layer, and regression layer, as illustrated in Figure \ref{fig-arch}. Our proposed architecture allows for the temporal representations learned by the LSTM to then be further learned for part-whole hierarchical relationships by the capsule attention through dynamic routing. Thus, capsule attention allows the model to learn which temporal representations to pay more attention to, given the uncertainties in the data as mentioned earlier in Section I.

\subsection{Input Representation Layer}
This layer encodes the input bio-signals as extracted fused features with three steps, namely data pre-processing, feature extraction, and feature fusion.

\subsubsection{Data Pre-processing}
Both EEG and EOG are first downsampled to $200$ \textit{Hz}, followed by a notch filter removing $50$ \textit{Hz} power line interference and a band-pass filter with a frequency range of $1.0-70$ \textit{Hz} minimizing artifacts such as noise \cite{zheng2017multimodal}. Min-max normalization of the signal amplitudes is employed to re-scale the biological time-series for each participant to $[-1, 1]$, thus minimizing the differences in signal amplitudes across different participants and signals \cite{wang2013real,segning2021neurophysiological,li2018exploring}.

\subsubsection{Feature Extraction}
\paragraph{EEG Features}
EEG signals are divided into non-overlapping $8$ second segments, where Short-time Fourier Transform (STFT) is used to calculate time-frequency features from $1$ second windows with $50\%$ overlap using a Hanning window. The log of the Power Spectral Density (PSD) and Differential Entropy (DE) are calculated on the STFT outputs with a $2$ \textit{Hz} resolution starting from $1.0$ \textit{Hz} \cite{zheng2017multimodal}. The PSD is calculated based on:
\begin{equation}
S_{xx}(\omega)=\lim_{T\to\infty}E\Big[|\hat{X}(\omega)|^{2}\Big], 
\end{equation}
and due to the Gaussian distribution of the signals, DE is calculated using:
\begin{equation}
DE = \frac{1}{2} \log{2\pi e \sigma^{2}}, \hspace{3mm} 
\end{equation}
where $x\sim N(\mu, \sigma^{2})$. 
Overall, we have $50$ EEG features extracted from each $1$-second window.

\paragraph{EOG Features}
We extract $36$ EOG features during three main type types of eye movements, namely blinking, saccade, and fixation, as described in \cite{zheng2017multimodal}. Blink and saccade movements can be detected from EOG recordings by applying peak detection on wavelet coefficients calculated with continuous wavelet transform \cite{zheng2017multimodal}. Statistical values such as mean, maximum, sum and variance are calculated during blink and saccades, as shown in Table \ref{table: eog feature}, where columns describe the features for each eye activity group, and the rows present the feature categories. We also compute blink numbers and saccade numbers. Other than features calculated during blink and saccade shown in the table, we also extract features during the fixation movement. Specifically, we calculate maximum and mean values of blink duration variance and saccade duration variance. We also measure the variance maximum, mean, and minimum of blink duration and saccade duration \cite{zheng2017multimodal}.

\begin{table}[!t]
\begin{center}
\centering\
\setlength\tabcolsep{3pt}
\caption{EOG features during blink and saccade.}
\label{table: eog feature}
% \small 
\scalebox{1} {
\begin{tabularx}{1\linewidth}{l|l|l}
	\hline
    &Blink &  Saccade\\
    \hline\hline
	\multirow{2}{*}{Rate}           & Maximum, Mean, Sum       & Maximum, Mean, Minimum        \\ 
	                                & Maximum, Mean of Variance    & Maximum, Mean of Variance \\ 
	\hline
	\multirow{3}{*}{Amplitude}      & Maximum, Minimum, Mean   & Maximum, Mean, Minimum        \\ 
	                                & Maximum, Mean of Variance    & Maximum, Mean of Variance  \\
	               	                & Power, Mean Power   & Power, Mean Power     \\ 
	\hline
% 	\hlines}

\end{tabularx}
}
\end{center}
\end{table}

\subsubsection{Feature Fusion}
EEG and EOG features are concatenated as follows: 
\begin{equation}
X_i^{Fused} = \big\{ X_i^{EEG} \cup {X_i^{EOG}}: \forall i \in [1, L] \big\}, 
\end{equation}
where $i$ denotes the $i$th feature sample and $L$ is the total number of feature samples for each participant. Accordingly, we have 
\begin{equation}
(X^{EEG}, X^{EOG}, X^{Fused}) \in (\mathbb{R}^{M\times L}, \mathbb{R}^{N\times L},\mathbb{R}^{(M+N)\times L}), 
\end{equation}
where $M$ and $N$ are the number of channels of the EEG and EOG signals respectively.

\subsection{Long Short-Term Memory Layer}
Our LSTM network \cite{greff2016lstm} employs a number of cells, the outputs of which are modified through the network by past information. Long-term dependencies are kept through the cells along the LSTM sequence using the common cell state. An input gate ($i_{t}$) and a forget gate ($f_{t}$) control the information flow and determine if the previous state ($C_{t-1}$) needs to be forgotten or if the current state ($C_t$) needs to be updated based on the latest inputs. An output gate ($o_{t}$) computes the output based on updated information from the cell state. Equations \ref{l1} through \ref{l5} present the LSTM cell architecture:
\begin{equation}\label{l1}
i_{t} = \sigma(W_{i}\cdot[h_{t-1},x_{t}]+b_{i}), 
\end{equation}
\begin{equation}\label{l2}
f_{t} = \sigma(W_{f}\cdot[h_{t-1},x_{t}]+b_{f}) ,
\end{equation}
\begin{equation}\label{l3}
C_{t}=f_t*C_{t-1} + i_t * tanh(W_c\cdot[h_{t-1}, x_{t}]+b_c) ,
\end{equation}
\begin{equation}\label{l4}
o_{t}=\sigma(W_{o}\cdot[h_{t-1},x_{t}]+b_{o}), 
\end{equation}
\begin{equation}\label{l5}
h_{t}=o_{t}*tanh(C_t) ,
\end{equation}
where $h_t$ and $h_{t-1}$ are the hidden states of the current and previous cells. $X_{t}$ is the cell input, $W_{f}, W_{i}, W_{c}, W_{o}$ are the weights, and $b_{f},b_{i},b_{c},b_{o}$ are the biases that we calculate and update using backpropagation along $t$.

\subsection{Feature Representation Layer}
This layer employs a lower level capsule layer and a higher level capsule layer to capture and cluster the representation of lower level features and higher level features with dynamic routing. 

\subsubsection{Lower Level Capsule Layer}
The output from each of the $L$ LSTM cells with $M$ hidden units is first reshaped as $[A_l \times A_w]$, where $A_l$ and $A_w$ define the grid of capsules. Then we split the LSTM cells into $C$ channels of $d-$dimensional capsules ($C \times d = L$), and within each, a convolution operation with an $e \times e$ kernel and stride of $g$ is employed. Accordingly, we produce $[C\times((A_l-e+1)/g\times (A_w-e+1)/g)]$ capsules where each contain a $d-$dimensional vector. Thus, each lower level capsule is represented as $u_i, i \in [1, C\times((A_l-e+1)/g\times (A_w-e+1)/g)].$

\subsubsection{Higher Level Capsule Layer}
This layer consists of a $K \times H$ matrix where $K$ is the number of higher level capsules and $H$ is the dimension of each higher level capsule $s_j, j \in [1, K]$. 

\subsubsection{Dynamic Routing}
The length of the higher level capsule output $v_j$ can be considered as the probability of existence of that higher level representation. Therefore, a non-linear squashing function $s_j$ 
% (see Eq. \ref{eq:squashing})
is employed to normalize $v_j$ into the range of $(0, 1)$ while the direction of $v_j$ remains unchanged. The squashing operation is performed as:
\begin{equation}\label{eq:squashing}
v_j = \frac{\|s_j\|^2}{1+{\|s_j\|}^{2}} \frac{s_j}{\|s_j\|},
\end{equation}
where $s_j$ is a weighted sum of $\hat{u}_{j|i}$ representing a prediction vector from lower level capsule $i$ to higher level capsule $j$ based on
\begin{equation}\label{eq:digital capsule}
s_j = \sum_i c_{ij}\hat{u}_{j|i},
\end{equation}
where $\hat{u}_{j|i}$ is calculated by the multiplication of a weight matrix $W_{ij}$ and a lower level capsule output $u_i$, where the size of $W_{ij}$ is $[d \times H]$. Therefore $\hat{u}_{j|i}$ is defined as
\begin{equation}\label{eq:prediction_vector}
\hat{u}_{j|i} = W_{ij} u_i.
\end{equation}

Coupling coefficients $c_{ij}$ between a lower level capsule $i$ and all the higher level capsules $j$ denote the probability of capsule $i$ being coupled to capsule $j$, where $c_{ij}$ is calculated using a softmax function for logit $b_{ij}$. Then, $b_{ij}$ are the log prior probabilities and $c_{ij}$ is therefore summed to $1$ by 
\begin{equation}
c_{ij} = \frac{exp(b_{ij})}{\sum_{j}exp(b_{ij})}. 
\end{equation}

Dynamic routing performs based on routing-by-agreement between $\hat{u}_{j|i}$ and $v_j$. The feature representation layer employs a dynamic routing algorithm \cite{sabour2017dynamic} to update zero initialized $b_{ij}$, by evaluating consistency between $\hat{u}_{j|i}$ and $v_j$ with an inner product $\hat{u}_{j|i} \cdot v_j$. Then $b_{ij}$ is updated to a higher value if $\hat{u}_{j|i}$ and $v_j$ have a strong agreement. Otherwise, a lower value is assigned to $b_{ij}$. To learn the part-whole relationships, Algorithm \ref{Dynamic Routing Algorithm} is used, where $\Omega_l$ denotes the set of capsules in layer $l$.

\begin{algorithm}[!t]
\caption{Dynamic Routing Algorithm} \label{Dynamic Routing Algorithm}
% \small
\begin{algorithmic}[1]
\Procedure {Routing}{$\hat{u}_{j|i}$, $r$, $l$}
\State Log prior probability initialization: $b_{ij}  \leftarrow 0 $
\For {r iterations} 
\ForAll {capsule $i \in \Omega_l$}

\State $c_i \leftarrow softmax(b_i) \hspace{20mm} $ 
\EndFor
\ForAll {capsule $j \in \Omega_{(l+1)}$}
\State $s_j \leftarrow \sum_{i} c_{ij}\hat{u}_{j|i}  \hspace{24mm} $
\State $v_j \leftarrow squash(s_j)  \hspace{23mm} $
\EndFor
\ForAll {capsule $i \in \Omega_l$, capsule $j \in \Omega(l+1)$}
\State $b_{ij} \leftarrow b_{ij} + \hat{u}_{j|i} \cdot v_j$
\EndFor
\EndFor
\EndProcedure
\end{algorithmic}
\end{algorithm}

\subsection{Regression Layer}
This layer contains a fully connected layer with a \textit{tanh} activation to ensure that the network predictions cover the range of recorded vigilance scores.

\section{Experiment Setup}
\label{sec:experi}
In order to evaluate the performance of our proposed solution for multimodal vigilance estimation, we conduct the following experiments.

\subsection{Dataset}
SEED-VIG is a large dataset for vigilance estimation where the data were collected from $23$ participants \cite{zheng2017multimodal}. Both EEG and EOG were collected using the ESI Neuroscan system\footnote{https://compumedicsneuroscan.com/} with a sampling rate of $1000$ \textit{Hz}. $17$ EEG channels were recorded from the temporal (`FT7', `FT8', `T7', `T8', `TP7', `TP8') and posterior (`CP1', `CP2', `P1', `PZ', `P2', `PO3', `POZ', `PO4', `O1', `OZ', `O2') brain regions and $4$ EOG channels were collected from the forehead. participants were required to drive the simulated car in a virtual environment for around $120$ minutes. Most of the participants were asked to perform the simulation after lunch to increase the possibility of fatigue \cite{zheng2017multimodal,ferrara2001much}. SMI eye-tracking glasses\footnote{https://www.smivision.com/eye-tracking/products/mobile-eye-tracking/} used an infrared camera to record eye gaze and several eye movements including blinks, eyes closures (CLOS), saccade, and fixation. Accordingly, vigilance score, PERCLOS \cite{dinges1998perclos}, is calculated as the percentage of blinks plus CLOS over the total duration of these four activities, described as:
\begin{equation}\label{PERCLOS}\small
    PERCLOS = \frac{blink+CLOS}{blink + fixation + saccade + CLOS}.
\end{equation}

\subsection{Implementation Details}
In our experiments, in order to solve the problem of different ranges and distribution of fused features, we employ a batch normalization layer \cite{ioffe2015batch} followed by a LeakyReLU \cite{maas2013rectifier} activation layer before each LSTM layer and lower level capsule layer, thus normalizing, re-scaling, and shifting the fused features. Batch normalization is not employed after the lower level capsule layer due to its negative effect on the squashing function. We employ Mean Square Error (MSE) $L = \frac{1}{N}\sum_{i=1}^{N}(y_i-\hat{y}_i)^2$ as the loss function and Adam optimizer \cite{kingma2014adam} to help minimize the loss. We use the default values of Adam optimizer \cite{kingma2014adam} and Batch normalization layers \cite{ioffe2015batch} to efficiently train our proposed model. We empirically tune the hyper-parameters of the network to achieve the best performance. The list of hyper-parameter settings is presented in Table \ref{eq_hyper}. The pipeline is implemented using TensorFlow \cite{abadi2016tensorflow} on a pair of NVIDIA RTX 2080Ti GPUs.

\begin{table}[!t]
\centering
\caption{Training hyper-parameters.}\label{eq_hyper}
\small
\scalebox{1} {
\begin{tabularx}{0.95\linewidth}{lll}
	\hline
% 	\hline
	%Paper & Method & Model Validation Methods & Average Accuracy\\
    {Layers} & {Parameters} &  {Value} \\
    \hline\hline
    \multirow{2}{*}{Model}      & Batch size          &    $32$  \\
                                & Training epochs     &    $30$ \\ 
    \hline
    % \hline
	\multirow{3}{*}{LSTM} & Recurrent depth     & {$3$} \\  
                          & Hidden layer units M & {$256$} \\ 
                          & No. of cells L      & {$15$} \\
    \hline
 	{Leaky ReLU} & Slope $\alpha$ & {$0.3$} \\   
 	\hline
	\multirow{5}{*}{Lower Level Caps}    & Kernel size e           & {$3$}    \\
                                         & Stride g           & {$1$}    \\	         
	                                     & No. of channels C       & {$5$}    \\ 
                                	     & Dimension size d        & {$3$}    \\ 
                                	     & Caps channel grid $[A_l, A_w]$ & {$[16, 16]$}   \\ 

     \hline                          	
   	\multirow{2}{*}{Higher Level Caps}   & No. of representations K  & {$10$} \\
                                	& Dimension size H  & {$16$}         \\ 
	\hline
	{Dynamic Routing} & Routing iterations r &$3$\\
	\hline
	{Regression} & Activation  & {tanh} \\ 

	\hline
% 	\hline
\end{tabularx}
}
\end{table}

\subsection{Evaluation Method}
To evaluate the performance of our regression method, the following two metrics are utilized:
\begin{equation}
    RMSE(Y, \hat{Y}) = \sqrt{\frac{1}{N}\sum_{i=1}^{N}{(y_i-\hat{y}_i)}^2} ,
\end{equation}
% \hspace{3}
\begin{equation}
    PCC(Y, \hat{Y}) = \frac{\sum_{i=1}^N(y_i-\Bar{y})(\hat{y}_i-\Bar{\hat{y}}))}{\sqrt{\sum_{i=1}^N(y_i-\Bar{y})^2\sum_{i=1}^N(\hat{y}_i-\Bar{\hat{y}})^2}},
\end{equation} where $Y$ is the vector of output PERCLOS labels and $\hat{Y}$ is the vector of predicted labels for all the samples. $y_i$ and $\hat{y_i}$ are the ground truth and prediction ratings for sample $i$, and $\Bar{y}$ and $\Bar{\hat{y}}$ are the mean ground truth and predicted ratings for all the samples.

We use both intra-participant and cross-participant validation schemes to evaluate the model performance in detail. We follow the same protocol as the works mentioned in the Related Works. We employ $5$-fold cross-validation method for the intra-participant scheme, where data for each participant are divided into $5$ folds. No overlap exists between the testing and training data. To perform cross-participant validation, we employ Leave-One-Participant-Out (also called Leave-One-Subject-Out (LOSO)) cross-validation, where the data from $22$ participants are used for training, and the remaining participant is used for testing. LOSO validation is critical in examining the participant-dependency of our method.

\subsection{Comparison}
\subsubsection{State-of-the-art methods} As described in the Related Works, a number of solutions have been proposed for this dataset. Here, we further describe the state-of-the-art solutions in both intra-participant and cross-participant validation scenarios. We strictly report related work that have used multi-modal approaches in order to provide a fair comparison.

\begin{itemize}
\item In \cite{huo2016driving}, GELM was employed by integrating graph regularization to ELM, thus establishing adjacent graph and constrain output weights by learning the similarity among the sample outputs and its $k$ nearest neighbors. Two fusion methods are proposed in order to achieve better performance in \textbf{intra-participant} validation. The feature level fusion helps the GELM model achieve the best performance.

\item In \cite{li2018multimodal}, two multimodal domain adaption networks notably DANN and ADDA based on feature fusion of EEG and EOG were proposed, optimizing transfer from data into the feature space. Both DANN and ADDA employed adversarial training to minimize prediction loss by eliminating domain shift between the source (training set) and target (testing set) domains. Feature level fusion is applied to obtain the best results for \textbf{cross-participant} estimation of vigilance scores.
\end{itemize}

\subsubsection{Baseline models} In addition to the methods published in the literature, we also implement four models for further benchmarking our proposed architecture. 

\begin{itemize}
\item  First, we utilize $3$ stacked 2D convolutional layers and $32$, $64$, and $128$ filters for the first, second, and third layers respectively. Each layer uses $3\times 3$ kernels and a stride of $1$ with ReLU activation. A fully connected layer with $512$ neurons has been applied after the last convolution layer to embed the feature maps.

\item Second, we utilize $3$ stacked LSTM layers, where each layer has $15$ cells and $256$ hidden units. The output has been computed from the last cell of the final LSTM layer. 

\item Third, we implement a cascade convolutional recurrent neural network (CNN-LSTM) by reproducing the same method used in \cite{zhang2018cascade}. We use the same setting as previous baseline CNN model. Then we implement stacked LSTM layers with the same settings as previous baseline LSTM model to further encode the embeddings in the previous fully connected layer.      

\item Lastly, we implement a capsule attention with the same parameters as our proposed model. 
\end{itemize}

For all these baseline methods, we implement a fully connected layer followed with a tanh activation function in order to perform the regression task. The parameters of all the baseline methods are tuned empirically to achieve the best results. Our implementation details and hyper-parameters in the baseline LSTM architecture (e.g. output layer activation function, number of LSTM units, optimizers, and training epochs) are different from \cite{zhang2016continuous}. Moreover, instead of dropout \cite{zhang2016continuous}, we employed batch normalization followed by a Leaky ReLU, which significantly improved the results.

\begin{table*}[t!]
\caption{The performance of our proposed model in comparison to different solutions using intra-participant validation.} \label{table: compare-intra}
\setlength\tabcolsep{20pt}
\centering
\label{table: intra-participant}
% \footnotesize
\small
% \resizebox{\textwidth}{!}
\begin{tabularx}{0.79\textwidth}{llll}
\hline
Paper & Method & RMSE$\pm$SD & PCC$\pm$SD \\
\hline
\hline
Zheng and Lu \cite{zheng2017multimodal} & SVR    & $0.10$    & $0.83 $\\
Zheng and Lu \cite{zheng2017multimodal} & CCRF   & $0.10$    & $0.84 $\\
Zheng and Lu \cite{zheng2017multimodal} & CCNF   & $0.09$    & $0.85 $\\
Du et al. \cite{du2017detecting}        & DAE    & $0.094  \pm 0.017$  & $0.852   \pm 0.064$\\
Huo et al. \cite{huo2016driving}        & GELM   & $0.0712$    & $0.8080$\\
Zhang et al. \cite{zhang2016continuous} & LSTM   & $0.0807 \pm0.0135$  & $0.8363  \pm 0.1009$\\
Wu et al. \cite{wu2018regression}       & DNNSN  & $0.08$    & $0.86 $\\
 \hline
Ours (baseline) & CNN & $0.0437 \pm 0.0079$ & $0.9320 \pm 0.0076$\\
Ours (baseline) & LSTM & $0.0477 \pm 0.0093$ & $0.8817 \pm 0.0083$\\
Ours (baseline) & CNN-LSTM & $0.0425 \pm 0.0091$ & $0.9387 \pm 0.0072$\\
Ours (baseline) & CapsNet & $0.0431 \pm 0.0069$ & $0.9381 \pm 0.0081$\\
\hline
\textbf{Ours} & \textbf{LSTM-CapsAtt} & $\mathbf{0.0295 \pm 0.0095}$ & $\mathbf{0.9887 \pm 0.0072}$\\
\hline
\end{tabularx}
\end{table*}

\begin{table*}[!t]%[!h]
\centering
\setlength\tabcolsep{20pt}
\caption{The performance of our proposed model in comparison to different solutions using cross-participant validation.} \label{table: compare-cross}
\label{table: cross-participant}
\small
\begin{tabularx}{0.8\textwidth}{llll}
	\hline
	Paper & Method & RMSE$\pm$SD & PCC$\pm$SD \\
	\hline
	\hline
	Li et al. \cite{li2018multimodal} & DANN      & $0.1427 \pm 0.0588$ & $0.8402 \pm0.1535$\\
	Li et al. \cite{li2018multimodal} & ADDA      & $0.1405 \pm 0.0514$ & $0.8442 \pm0.1336$\\
 	\hline
 	Ours (baseline)         & CNN       & $0.1341 \pm 0.0919$ & $0.8491 \pm 0.1472$\\
    Ours (baseline)         & LSTM      & $0.1321 \pm 0.0981$ & $0.8537 \pm 0.1476$\\
    Ours (baseline)         & CNN-LSTM  & $0.1297 \pm 0.0750$ & $0.8603 \pm 0.1277$\\
    Ours (baseline)         & CapsNet   & $0.1317 \pm 0.0933$ & $0.8587 \pm 0.1177$\\
    \hline
    \textbf{Ours} & \textbf{LSTM-CapsAtt} & $\mathbf{0.1089 \pm 0.0696}$ & $\mathbf{0.8823 \pm 0.1084}$\\
    \hline
\end{tabularx}
\end{table*}

\section{Results and Discussion} 
In this section, we present the results of our proposed architecture and compare the performance to other published solutions, as well as a number of baseline methods, in both intra-participant and cross-participant schemes. Additionally, we perform extensive ablation experiments to investigate the effects of different variations in the model architecture, routing iterations, different attention mechanisms, and different modalities. Next, we analyze the topological patterns of correlation between EEG and output vigilance scores. We further discuss the role of capsule attention in addressing the aforementioned challenges by analyzing the embeddings obtained from different levels of capsules. Lastly, we investigate the impact of noise on our capsule-based model and visualize the embeddings obtained by our model from different modalities.

\subsection{Performance}
Tables \ref{table: intra-participant} and \ref{table: cross-participant} present the performance of our proposed architecture in comparison to the other aforementioned methods for both validation scenarios. The evaluation metrics RMSE and PCC listed in the tables are achieved using multimodal EEG and EOG. It is observed that the LSTM-CapsAtt model achieves state-of-the-art results by outperforming both previous solutions and baseline methods, based on both RMSE and PCC values.
This confirms that the obtained embeddings in the high-level capsule layer (see Figure \ref{fig-arch}) are informative for multimodal vigilance estimation. Since the improvement in the cross-participant validation scheme is larger than in intra-participant validation, it can be concluded that the representations obtained through our capsule attention mechanism are more discriminative for learning high-level participant-independent attributes, contributing to the more difficult task of cross-participant validation.

\subsection{Ablation Experiments}
\subsubsection{Effect of LSTM Architecture}
Here, we evaluate the effect of several important parameters, notably the number of stacked LSTM layers and activation function used for the regression layer on the results. The performances are outlined in Table \ref{table:variants-intra} and Table \ref{table:variants-cross} for intra-participant and cross-participant validations respectively. The results show that three stacked LSTM layers help our model achieve the best results in both validation scenarios. Activation functions also play a critical role in the regression model, where tanh outperforms Sigmoid and ReLU activation functions for the proposed model in all the scenarios with different stacked LSTM layers. ReLU performs poorly in the proposed model mainly due to the lack of constraint on the model output.

\begin{table}[!htp]
\begin{center}
\centering
\caption{Comparison of our proposed model with different variants using intra-participant validation.}\label{table:variants-intra}
\small
\scalebox{1} {
\begin{tabularx}{0.97\columnwidth}{l|l|l|l}
	\hline
    Layers & Activation & RMSE & PCC \\
    \hline\hline
    1 & ReLU & $0.1539 \pm 0.0139$   & $0.7828 \pm 0.0737$ \\ 
	2 & ReLU & $0.1515 \pm 0.19$   & $0.7830 \pm 0.0739$ \\ 
	3 & ReLU & $0.1303 \pm 0.0136$   & $0.7857 \pm 0.0727$ \\
	4 & ReLU & $0.1305 \pm 0.0130$   & $0.7855 \pm 0.0729$ \\
	5 & ReLU & $0.1305 \pm 0.0131$   & $0.7851 \pm 0.0727$ \\ \hline
    1 & Sigmoid & $0.0352 \pm 0.0107$   & $0.9861 \pm 0.0079$ \\
	2 & Sigmoid & $0.0325 \pm 0.0092$   & $0.9870 \pm 0.0082$ \\   
	3 & Sigmoid & $0.0324 \pm 0.0092$   & $0.9870 \pm 0.0083$ \\ 
	4 & Sigmoid & $0.0325 \pm 0.0091$   & $0.9869 \pm 0.0084$ \\
	5 & Sigmoid & $0.0324 \pm 0.0091$   & $0.9869 \pm 0.0083$ \\ \hline
	1 & tanh & $0.0322 \pm 0.0097$   & $0.9873 \pm 0.0081$ \\ 
	2 & tanh & $0.0313 \pm 0.0095$   & $0.9882 \pm 0.0075$ \\ 
	3 & tanh & $\mathbf{0.0295 \pm 0.0095}$   & $\mathbf{0.9887 \pm 0.0072}$ \\ 
	4 & tanh & $0.0321 \pm 0.0099$   & $0.9875 \pm 0.0080$ \\
	5 & tanh & $0.0315 \pm 0.0096$   & $0.9880 \pm 0.0075$ \\
	\hline
% 	\hline
\end{tabularx}
}
\end{center}
\end{table}

\begin{table}[!htp]
\begin{center}
\centering
\caption{Comparison of our proposed model with different variants using cross-participant validation.}\label{table:variants-cross}
\small 
\scalebox{1} {
\begin{tabularx}{0.97\columnwidth}{l|l|l|l}
	\hline
    Layers & Activation & RMSE & PCC \\
    \hline\hline
	1 & ReLU & $0.2027 \pm 0.1150$   & $0.8026 \pm 0.1887$ \\ 
	2 & ReLU & $0.2057 \pm 0.1131$   & $0.8002 \pm 0.1956$ \\ 
	3 & ReLU & $0.1854 \pm 0.1011$   & $0.8079 \pm 0.1833$ \\
	4 & ReLU & $0.1899 \pm 0.1037$   & $0.8059 \pm 0.1877$ \\
	5 & ReLU & $0.1897 \pm 0.1033$   & $0.8061 \pm 0.1865$ \\ \hline
    1 & Sigmoid & $0.1116 \pm 0.0745$   & $0.8749 \pm 0.1113$ \\ 
	2 & Sigmoid & $0.1098 \pm 0.0699$   & $0.8813 \pm 0.1083$ \\ 
	3 & Sigmoid & $0.1093 \pm 0.0667$   & $0.8820 \pm 0.1088$ \\ 
	4 & Sigmoid & $0.1106 \pm 0.0671$   & $0.8810 \pm 0.1085$ \\
	5 & Sigmoid & $0.1105 \pm 0.0671$   & $0.8813 \pm 0.1080$ \\ \hline
	1 & tanh & $0.1112 \pm 0.0703$   & $0.8797 \pm 0.1089$ \\ 
	2 & tanh & $0.1101 \pm 0.0654$   & $0.8810 \pm 0.1080$ \\ 
	3 & tanh & $\mathbf{0.1089 \pm 0.0696}$   & $\mathbf{0.8823 \pm 0.1084}$ \\ 
	4 & tanh & $0.1103 \pm 0.0657$   & $0.8807 \pm 0.1082$ \\
	5 & tanh & $0.1101 \pm 0.0661$   & $0.8811 \pm 0.1080$ \\

	\hline
% 	\hline
\end{tabularx}
}
\end{center}
\end{table}

\subsubsection{Effect of Routing Iterations} % Should be Capitalized 
\begin{figure}[!t]
% \label{fig:iterations}
    \begin{center}
    \includegraphics[width=0.85\columnwidth]{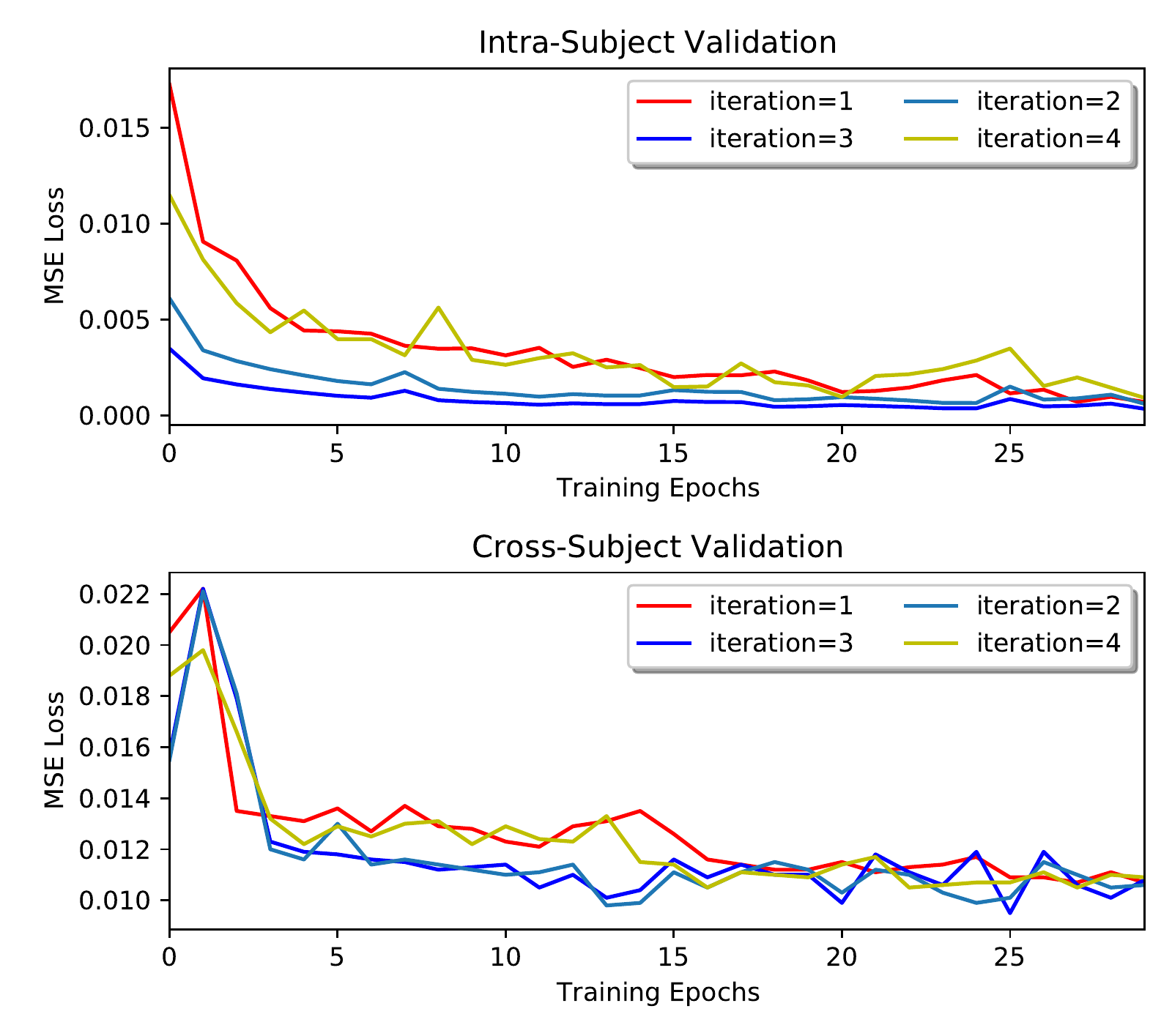}
    \end{center}
\caption{Effect of routing iterations.}
\label{fig:iterations}
\end{figure}

To investigate the effect of routing iterations on our proposed model, we conduct experiments with different numbers of routing iterations using both validation scenarios. Figure \ref{fig:iterations} shows the calculated MSE loss of the model for 30 training epochs. The model achieves the best results with 3 iterations, showing fast convergence of the dynamic routing algorithm in conformity with \cite{sabour2017dynamic}.

\subsubsection{Effect of Attention Mechanisms}
We evaluate different attention mechanisms in comparison to our proposed model. To this end, we employ LSTM-CNN and LSTM-SoftAtt architectures using the same LSTM settings. The LSTM-CNN architecture employs CNN-based attention with the same parameters as the baseline CNN model and the LSTM-SoftAtt model employs the same soft attention mechanism as described in \cite{wang2016attention}. All the above-mentioned models have the same fully-connected layer with tanh activation. These settings were selected to maximize performance. As shown in Tables \ref{table:attention-intra} and \ref{table:attention-cross}, our approach outperforms the other solutions by achieving the best RMSE and PCC values in both validation scenarios.

\begin{table}[!t]
\begin{center}
\centering
\caption{Comparison of our model with other attention mechanisms using intra-participant validation.}\label{table:attention-intra}
% \small 
\scalebox{1} {
\begin{tabularx}{1\linewidth}{l|l|l}
	\hline
    Model &  RMSE & PCC \\
    \hline\hline
	LSTM-SoftAtt                  & $0.0397 \pm 0.0115$   & $0.9696 \pm 0.0092$  \\ 
	LSTM-CNN                        & $0.0424 \pm 0.0122$   & $0.9592 \pm 0.0087$  \\ 
	LSTM-CapsAtt (ours)             & $\mathbf{0.0295 \pm 0.0095}$  & $\mathbf{0.9887 \pm 0.0072}$ \\
	\hline
% 	\hline
\end{tabularx}
}
\end{center}
\end{table}

\begin{table}[!t]
\begin{center}
\centering
\caption{Comparison of our model with other attention mechanisms using cross-participant validation.}\label{table:attention-cross}
% \small 
\scalebox{1} {
\begin{tabularx}{1\linewidth}{l|l|l}
	\hline
    Model &  RMSE & PCC \\
    \hline\hline
	LSTM-SoftAtt                  & $0.1230 \pm 0.0712$   & $0.8688 \pm 0.1137$  \\ 
	LSTM-CNN                        & $0.1298 \pm 0.0759$   & $0.8592 \pm 0.1283$  \\ 
	LSTM-CapsAtt (ours)             & $\mathbf{0.1089 \pm 0.0696}$   & $\mathbf{0.8823 \pm 0.1084}$ \\ 
	\hline
% 	\hlines}

\end{tabularx}
}
\end{center}
\end{table}

\subsubsection{Effect of Multimodality}
To evaluate the effect of multimodality of EEG and EOG representations on the model performance, we compare the results of using single modalities (EEG alone and EOG alone) to the results of our multimodal solution. The results are presented in Table \ref{table:modality-intra} and Table \ref{table:modality-cross} for intra-participant and cross-participant validation scenarios respectively. For single modality, EEG slightly outperforms EOG in intra-participant validation, while EOG provides better performance in cross-participant validation. This observation could indicate that the inter-participant variability of EEG recordings is higher than that of EOG. However, the proposed multimodal EEG-EOG achieves superior performance to both unimodal EEG and unimodal EOG, indicating that the two modalities are likely to contain complimentary information about vigilance.

\begin{table}[!t]
\begin{center}
\centering
\setlength\tabcolsep{15 pt}
\caption{Effect of multimodality using intra-participant validation.}\label{table:modality-intra}
% \small 
% \scalebox{1} {

\begin{tabularx}{1.0\columnwidth}{c|c|c}
	\hline
	Modality       & RMSE & PCC \\ 
	\hline\hline
	EEG            & $0.0370\pm0.0117$  & $0.9852\pm0.0072$ \\ 
	EOG            & $0.0485\pm0.0126$  & $0.9709\pm0.0236$ \\ 
	EEG-EOG        & $0.0295\pm0.0095$  & $0.9887\pm0.0072$ \\ 
	\hline
% 	\hlines}

\end{tabularx}
% }
\end{center}
\end{table}

\begin{table}[!t]
\begin{center}
\centering
\setlength\tabcolsep{15 pt}
\caption{Effect of multimodality using cross-participant validation.}
\label{table:modality-cross}
% \small 
% \scalebox{1} {

\begin{tabularx}{\columnwidth}{c|c|c}
	\hline
	Modality       & RMSE & PCC \\ 
	\hline\hline
	EEG            & $0.1771\pm0.1030$ & $0.6713\pm0.2052$ \\ 
	EOG            & $0.1241\pm0.0713$ & $0.8393\pm0.1500$ \\ 
	EEG-EOG        & $0.1089\pm0.0696$ & $0.8823\pm0.1084$ \\ 
	\hline
% 	\hlines}

\end{tabularx}
% }
\end{center}
\end{table}

\subsection{Discussion}
\subsubsection{Topological Analysis}

\begin{figure}[!htb]
    \begin{center}
    \includegraphics[width=1\linewidth]{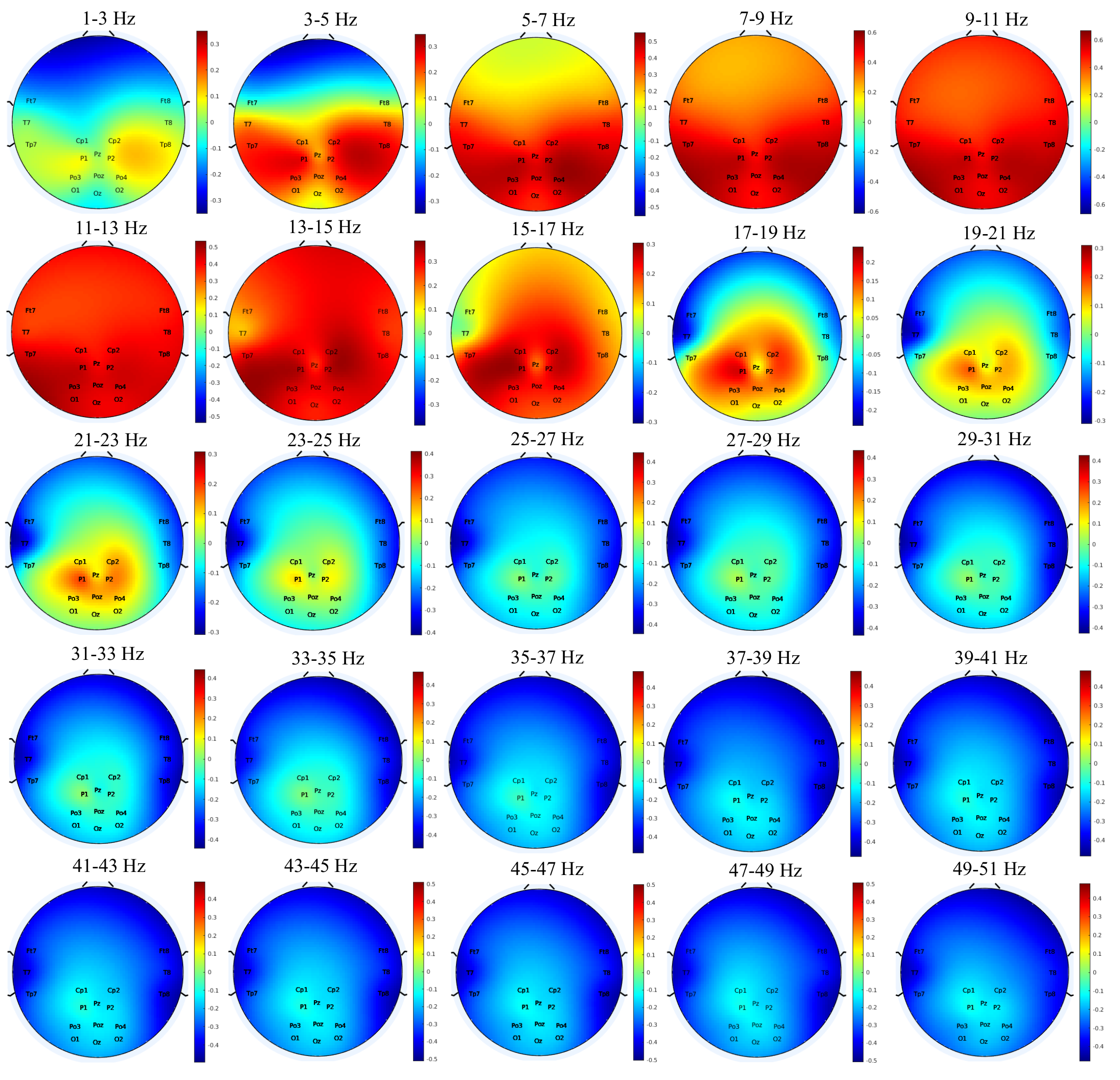} 
    \end{center}
\caption{Average PCC values between DE features and PERCLOS labels across all the participants, presented as topology graphs.}
\label{fig: topo}
\end{figure}

\begin{figure}[!ht]
    \begin{center}
    \includegraphics[width=1.0\columnwidth]{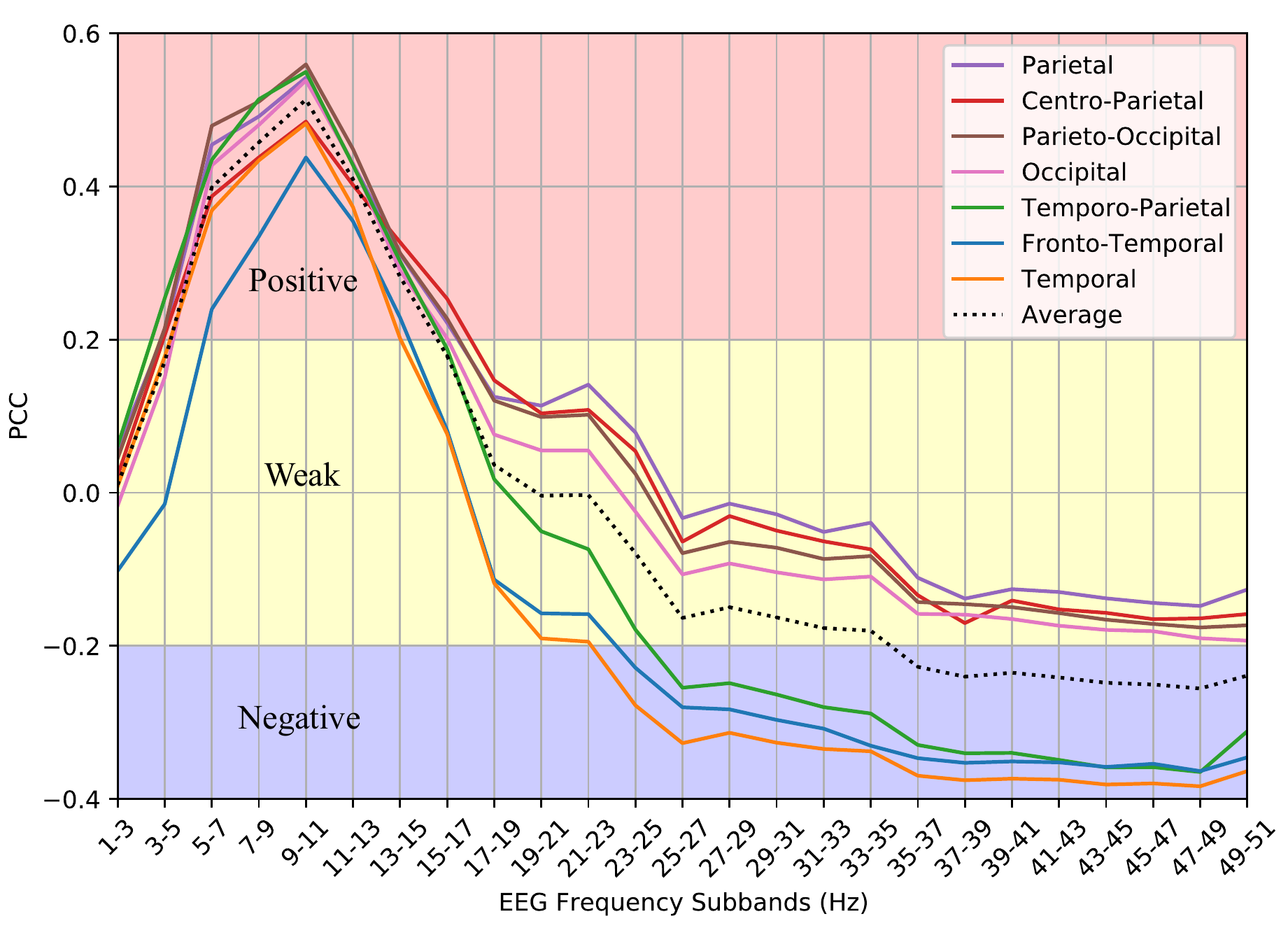} 
    \end{center}
\caption{{PCC patterns of different brain regions in the range of $1-51$ \textit{Hz}}.}
\label{fig: topo-analysis}
\end{figure}

We investigate topological patterns of correlation (PCC values) calculated between EEG features (differential entropy) and continuous PERCLOS values. We use differential entropy as prior works have shown its effectiveness in measuring the complexity and activity of EEG signals \cite{shi2013differential,zheng2017multimodal}.
Both EEG features and PERCLOS values are calculated for each $8$-second EEG segment. Accordingly, the data collected from the $23$ participants consists of $885$ consecutive EEG segments along with the corresponding continuous (in the range of $0$ to $1$) PERCLOS labels. Here, we calculate the PCC values for each of the $23$ participants, and visualize the average results on topology graphs in Figure \ref{fig: topo} for different EEG frequency sub-bands. We also demonstrate the correlation patterns of different brain regions in different frequency sub-bands in Figure \ref{fig: topo-analysis}. In this figure, the dotted line denotes the average value for all the regions. Accordingly, we specify three ranges of PCC values, namely \textit{positive}, \textit{weak}, and \textit{negative}. Based on this analysis, we observe that frequency bands in the range of $1-5$ \textit{Hz} as well as $17-35$ \textit{Hz} show weak correlations, while frequency bands in the range of $5-17$ \textit{Hz} exhibit strong positive correlations, and frequency bands in the range of $35-51$ \textit{Hz} show strong negative correlations. One-way analysis of variance (ANOVA) shows that the positive range ($M=0.38$, $SD=0.05$) is significantly higher than the weak range ($M=-0.05$, $SD=0.10$) with $F(17)=225.16$, $p = 4.92e^{-16}$, indicating high correspondence between PERCLOS values (used to quantify vigilance) and input EEG. This finding is consistent with \cite{torsvall1987sleepiness} where it was shown that alpha waves ($8-13$ \textit{Hz}) are effective in detecting drowsiness. However, our study finds that in addition to $8-13$ \textit{Hz}, frequency bands in the range of $5-8$ \textit{Hz} and $13-17$ \textit{Hz} also contain valuable information for vigilance estimation. Moreover, our analysis shows that the negative range ($M=-0.23$, $SD=0.10$) is significantly lower than the weak range ($M=-0.05$, $SD=0.10$) with $F(17)=24.61$, $p = 2.22e^{-5}$, indicating that frequency bands in the region of $35-51$ \textit{Hz} can also be effective for vigilance estimation.

Further analyzing the patterns in Figures \ref{fig: topo} and \ref{fig: topo-analysis}, we observe that in the negative range ($35-51$ \textit{Hz}), the fronto-temporal ($M=-0.35$, $SD=0.02$), temporal ($M=-0.37$, $SD=0.01$), and temporo-parietal ($M=-0.34$, $SD=0.05$) regions show significantly less correlation with vigilance than the centro-parietal ($M=-0.15$, $SD=0.02$), parietal ($M=-0.13$, $SD=0.02$), parieto-occipital ($M=-0.16$, $SD=0.02$) and occipital ($M=-0.17$, $SD=0.01$) regions. The $F$ and $p$ values obtained by ANOVA are presented in Table \ref{table:anova}. This finding suggests that in the range of $35-51$ \textit{Hz}, the fronto-temporal, temporal, and temporo-parietal regions are better indicators of driver vigilance in comparison to the other regions.

\begin{table*}[!t]
\begin{center}
\scriptsize
\centering
\setlength\tabcolsep{3 pt}
\caption{{ANOVA results indicating significant difference between the PCC values for different brain regions in the range of $35-51$ \textit{Hz}. FT: Fronto-Temporal , T: Temporal, TP: Temporo-Parietal, CP: Centro-Parietal, P: Parietal, PO: Parieto-Occipital, O: Occipital.}}
\label{table:anova}

% \scalebox{1} {

\begin{tabularx}{2.0\columnwidth}{c|c|c|c|c|c|c|c|c|c|c|c|c}
	\hline
	Brain Region 1       & FT & FT & FT & FT   & T & T& T & T      & TP & TP& TP & TP \\ 
	\hline
	Brain Region 2       & CP & P & PO & O     & CP & P & PO &O    & CP & P & PO & O  \\ 
	\hline\hline
	
	$F$(8)        & $1652.44$      & $2232.75$       & $1430.26$      & $1197.72$        & $1965.16$      & $2591.90$         & $1715.60$      & $1462.23$       & $610.52$     & $791.50$       & $556.85$      & $467.38$  \\ 
	\hline
	$p$           & $6.20e^{-16}$  & $7.66e^{-17}$   & $1.69e^{-15}$  & $5.78e^{-15}$   & $1.86e^{-16}$   & $2.71e^{-17}$     & $4.78e^{-16}$  & $1.45e^{-15}$   & $6.01e^{-13}$ & $1.01e^{-13}$ & $1.12e^{-12}$ & $3.73e^{-12}$  \\ 
	\hline
% 	\hlines}

\end{tabularx}
% }
\end{center}
\end{table*}

\begin{figure}[!t]
    \begin{center}
    \includegraphics[width=1\columnwidth]{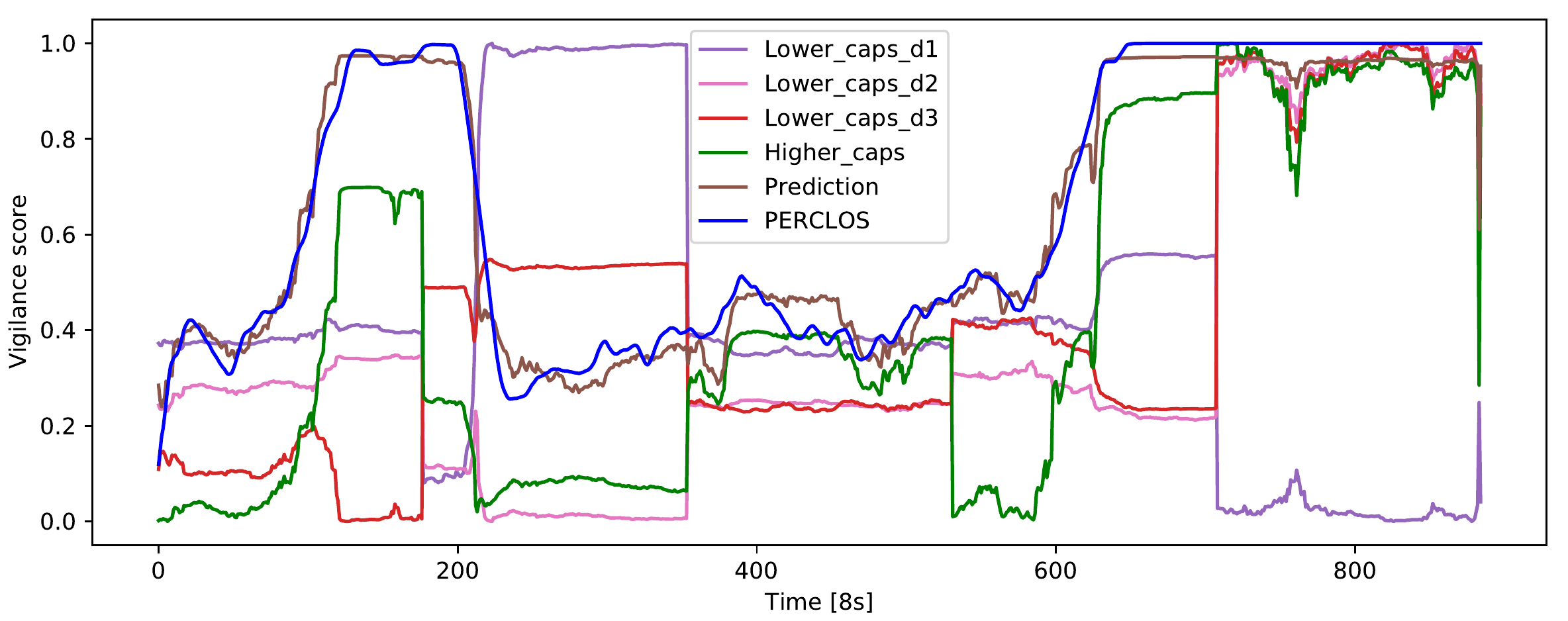} 
    \end{center}
\caption{Continuous vigilance estimation using embeddings from different level of capsules in one experiment.}
\label{fig: capsule-analysis}
\end{figure}

\subsubsection{Role of Capsule Attention}
To investigate the importance of capsule attention in our architecture for vigilance estimation, we compare the features embedded in lower level capsules and higher level capsules. First, we obtain the embedding from each dimension of lower level capsules ($u_i$). Next, we capture the embedding of squashed higher level capsules ($v_j$). We then use PCA to reduce these embeddings to one dimension. We further normalize the embeddings to $[0,1]$ and compare them with the output PERCLOS. As shown in Figure \ref{fig: capsule-analysis}, embeddings in each dimension of lower level capsules (presented as `lower$\_$caps$\_$d$1$', `lower$\_$caps$\_$d$2$', and `lower$\_$caps$\_$d$3$') sometimes behave similar, and sometimes very different with respect to one another. This phenomenon is mainly due to the aforementioned challenge that complementary and contradicting features of EEG and EOG signals existed in the embeddings of different dimensions of lower level capsules. In order to address this challenge, we learn the part-whole relationship among lower level capsules and higher level capsules. Specifically, we use capsule attention to assign lower level capsules ($u_i$) to higher level capsules ($s_j$) using the dynamic routing algorithm. The noise and artifacts caused by motion and muscle activities that are reflected in lower level capsules are reduced in higher level capsules (denoted as `Higher$\_$caps') since the embedding of ($s_j$) is much more correlated with PERCLOS. Therefore, capsule attention addresses the aforementioned challenges by \textit{i}) identifying the complementary and contradicting among embeddings in different dimensions of lower level capsules and \textit{ii}) reducing artifacts and noise existing in lower level capsules. Overall, our model provides a promising prediction in vigilance estimation (denoted as `Prediction'). Lastly, the capsule attention mechanism cannot be replaced by soft attention since the latter method is not able to assign attention scores from lower level embeddings to higher level embeddings.

\begin{figure}[!t]
    \begin{center}
    \includegraphics[width=1.0\linewidth]{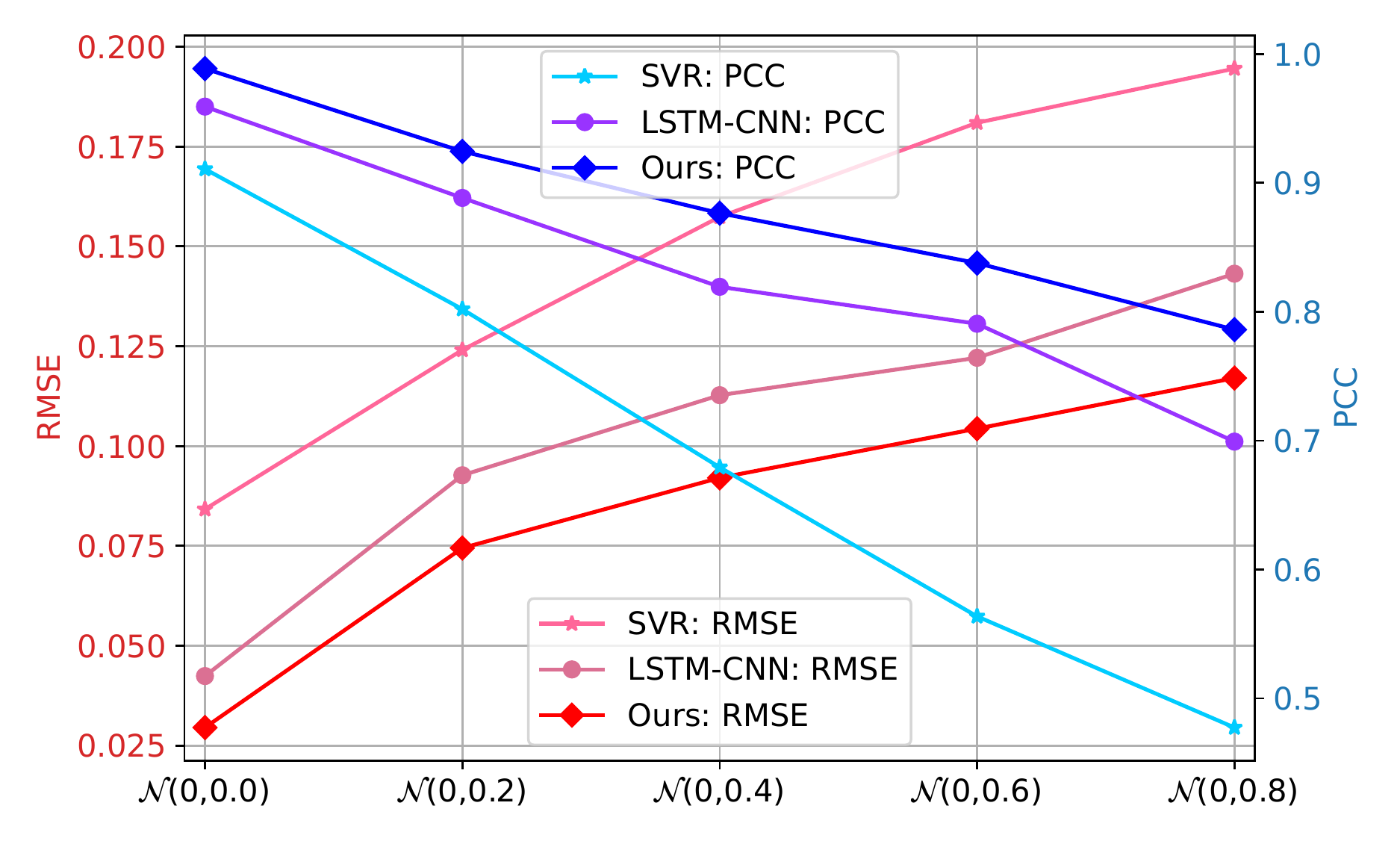} 
    \end{center}
\caption{{Performance when different amounts of Gaussian noise are added to input features.}}
\label{fig: noise}
\end{figure}

\subsubsection{Robustness to Noise}
To evaluate the impact of noise on our model, we perform additional experiments, where we compare the results obtained by our proposed architecture (LSTM-CapsAtt) in comparison to two baselines (SVR and LSTM-CNN) when different amounts of Gaussian noise are added to the input feature representations. The parameters of the LSTM-CNN are the same as the model described in the ablation experiments (Section V.B.3), while the parameters of the SVR are chosen via a grid search to obtain the best performance. The results are presented in Figure \ref{fig: noise}. We observe that the performance of all three models naturally decrease when the standard deviation of the Gaussian noise is increased ($\mathcal{N}(0, 0.2)$, $\mathcal{N}(0, 0.4)$, $\mathcal{N}(0, 0.6)$, and $\mathcal{N}(0, 0.8)$), with SVR showing the worst overall performance and our method showing the best performance. Moreover, we observe that as the noise intensifies, our model experiences a smaller drop in performance in comparison to SVR and LSTM-CNN. This added robustness to noise could be due to the fact that capsule attention employs routing by agreement to enable the lower level capsules to learn what needs to be paid attention to given the feedback from higher level capsules. Accordingly, when a lower level capsule contains noise, it is likely to make predictions that are far from the predictions made by other lower level capsules. However, routing by agreement which is used in the capsule network will reduce the contribution of that capsule, resulting in noise suppression in the network output \cite{hahn2019self}. On the other hand, as confirmed by the results in Figure \ref{fig: noise}, CNNs generally lack such mechanisms to reduce noise.

\subsubsection{Impact of Multimodality}
\begin{figure}[!t]
    \begin{center}
    \includegraphics[width=1\linewidth]{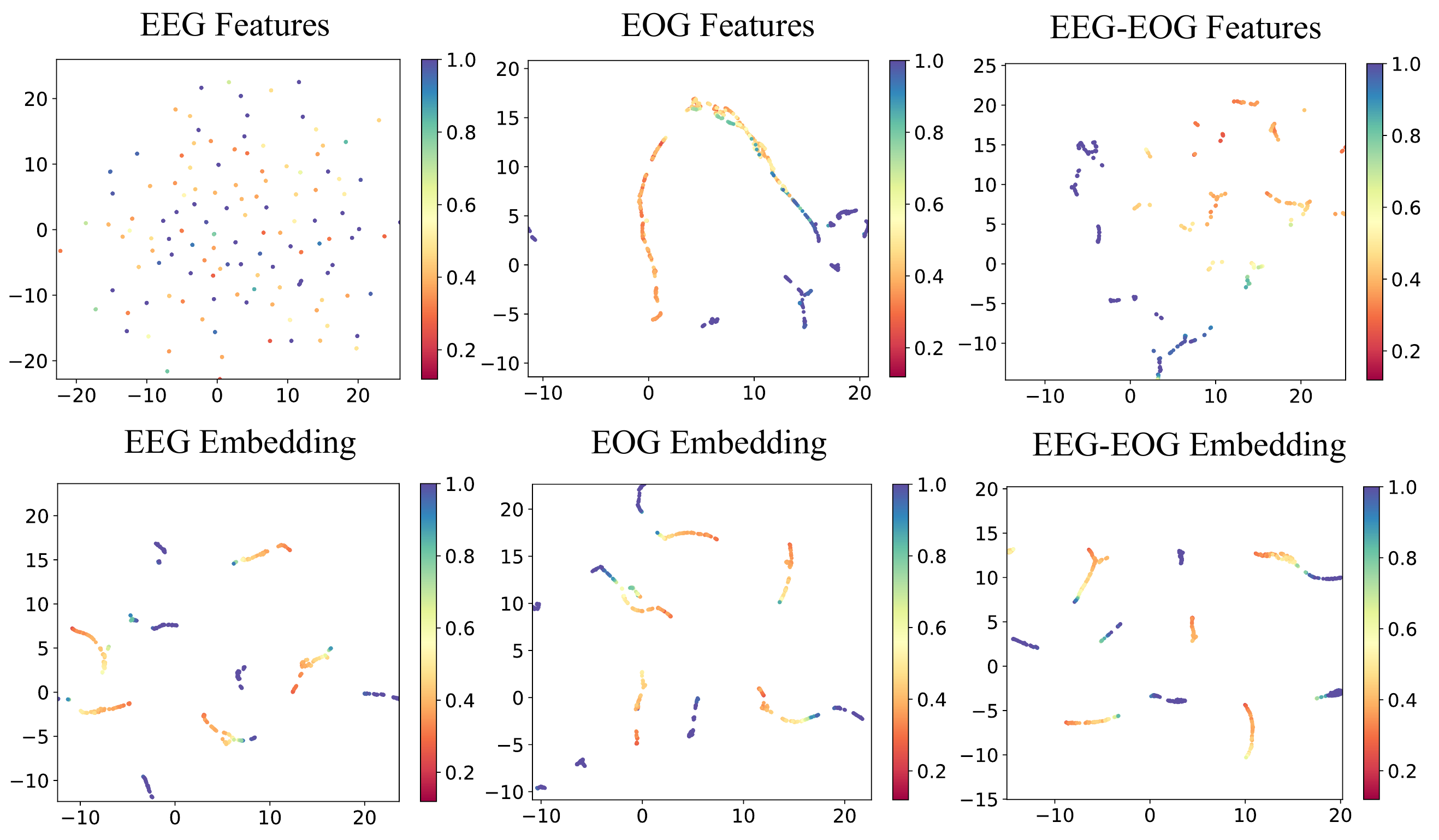} 
    \end{center}
\caption{Comparison between feature vectors ($1^{st}$ row) and feature embeddings obtained in our proposed network ($2^{nd}$ row) using UMAP.}
\label{fig: umap}
\end{figure}

We utilize Uniform Manifold Approximation and Projection (UMAP) \cite{mcinnes2018umap} to visualize the learned representations in order to better understand the effect of multimodality in our architecture. Figure \ref{fig: umap} shows the comparison of the features without using our architecture and the embeddings obtained from our proposed network using unimodal EEG, unimodal EOG, and multimodal EEG-EOG as inputs. We obtain the embedding space from the output of the squashed higher level capsule layer ($v_j$), as shown in Figure \ref{fig-overview}. We observe that the embeddings extracted from $v_j$ are more separable compared to the features without our proposed network in both unimodal and multimodal representations learning.

\section{Conclusions}
\label{sec:conclu}
In this paper, we propose a novel multimodal approach based on an architecture consisting of an LSTM network with Capsule Attention for in-vehicle vigilance estimation from EEG and EOG. Our model extracts lower level hierarchical information using a lower level capsule layer and further captures and groups these representations with a higher level capsule layer, where part-whole relationships in the features are explored using dynamic routing. The experiments show the generalizability of our model by achieving state-of-the-art results in both intra-{participant} and cross-{participant} validation scenarios. The results confirm the impact of capsule attention on multimodal EEG-EOG representation learning for in-vehicle driver vigilance estimation, and its ability to deal with noise better than other baselines.
\label{sec:refs}

\bibliographystyle{IEEEtran}
\bibliography{IEEEabrv,refs}

\end{document}